\documentclass{article}

\usepackage{microtype}
\usepackage{graphicx}
\usepackage{booktabs} 
\usepackage{float}

\usepackage{hyperref}

\usepackage{graphicx}
\usepackage{subcaption}
\usepackage{threeparttable}

\usepackage{amsmath}
\usepackage{amssymb}
\usepackage{amsfonts}
\usepackage{amsthm}
\usepackage{multirow}

\newtheorem{definition}{Definition}

\usepackage[accepted]{icml2018}

\icmltitlerunning{Dimensionality-Driven Learning with Noisy Labels}

\begin{document}

\twocolumn[
\icmltitle{Dimensionality-Driven Learning with Noisy Labels}



\icmlsetsymbol{equal}{*}

\begin{icmlauthorlist}
\icmlauthor{Xingjun Ma}{equal,uom}
\icmlauthor{Yisen Wang}{equal,tsinghua}
\icmlauthor{Michael E. Houle}{nii}
\icmlauthor{Shuo Zhou}{uom}
\icmlauthor{Sarah M. Erfani}{uom}
\icmlauthor{Shu-Tao Xia}{tsinghua}
\icmlauthor{Sudanthi Wijewickrema}{uom}
\icmlauthor{James Bailey}{uom}
\end{icmlauthorlist}

\icmlaffiliation{uom}{The University of Melbourne, Melbourne, Australia}
\icmlaffiliation{tsinghua}{Tsinghua University, Beijing, China}
\icmlaffiliation{nii}{National Institute of Informatics, Tokyo, Japan}

\icmlcorrespondingauthor{Yisen Wang}{wangys14@mails.tsinghua.edu.cn}
\icmlcorrespondingauthor{Xingjun Ma}{xingjun.ma@unimelb.edu.au}

\icmlkeywords{Deep Neural Networks, Local Intrinsic Dimensionality, Generalization, Noisy Labels}

\vskip 0.3in
]



\printAffiliationsAndNotice{\icmlEqualContribution} 

\begin{abstract}
Datasets with significant proportions of noisy (incorrect) class labels present challenges for training accurate Deep Neural Networks (DNNs).  We propose a new perspective for understanding DNN generalization for such datasets, by investigating the dimensionality of the deep representation subspace of training samples.  We show that from a dimensionality perspective, DNNs exhibit quite distinctive learning styles when trained with clean labels versus when trained with a proportion of noisy labels.
Based on this finding, we develop a new dimensionality-driven learning strategy, which monitors the dimensionality of subspaces during training and adapts the loss function accordingly.
We empirically demonstrate that our approach is highly tolerant to significant proportions of noisy labels, and can effectively learn low-dimensional local subspaces that capture the data distribution. 

\end{abstract}

\section{Introduction}
\label{sec:intro}
Deep Neural Networks (DNNs) have demonstrated excellent performance in solving many complex problems, and have been widely employed for tasks such as speech recognition \cite{hinton2012deep}, computer vision \cite{he2016deep} and gaming agents \cite{silver2016mastering}. DNNs are capable of learning very complex functions, and can generalize well even for a huge number of parameters \cite{neyshabur2014search}. However, recent studies have shown that DNNs may generalize poorly for datasets which contain a high proportion noisy (incorrect) class labels  \cite{zhang2016understanding}. It is important to gain a fuller understanding of this phenomenon, with a view to development of new training methods that can achieve good generalization performance in the presence of variable amounts of label noise.

One simple approach for noisy labels is to ask a domain expert to relabel or remove suspect samples in a preprocessing stage. However, this is infeasible for large datasets and also runs the risk of removing crucial samples.  An alternative is to correct noisy
labels to their true labels via a clean label inference step \cite{vahdat2017toward,veit2017learning,jiang2017mentornet,li2017learning}. Such methods often assume the availability of a supplementary labelled dataset containing pre-identified noisy labels which are used to develop a model of the label noise. However, their effectiveness is tied to the assumption that the data follow the noise model. A different approach to tackle noisy labels is to utilize correction methods such as loss correction \cite{patrini2017making,ghosh2017robust}, label correction \cite{reed2014training}, or additional linear correction layers \cite{sukhbaatar2014learning,goldberger2016training}.

In this paper, we first investigate the dimensionality of the deep representation subspaces learned by a DNN and provide a dimensionality-driven explanation of DNN generalization behavior in the presence of (class) label noise. Our analysis employs a dimensionality measure called Local Intrinsic Dimensionality (LID) \cite{houle13,houle2017local1}, applied to the deep representation subspaces of training examples. 
We show that DNNs follow two-stage of learning in this scenario: 1) an early stage of {\em dimensionality compression}, that models low-dimensional subspaces that closely match the underlying data distribution, and 2) a later stage of {\em dimensionality expansion}, that steadily increases subspace dimensionality in order to overfit noisy labels. This second stage appears to be a key factor behind the poor generalization performance of DNNs for noisy labels. Based on this finding, we propose a new training strategy, termed {\em Dimensionality-Driven Learning}, 
that avoids the dimensionality expansion stage of learning by adapting the loss function.   Our main contributions are: 

\begin{itemize}
  \item We show that from a dimensionality perspective, DNNs exhibit distinctive learning styles with clean labels versus noisy labels.
  
  \item We show that the local intrinsic dimensionality can be used to identify the stage shift from dimensionality compression to dimensionality expansion.
  
  \item We propose a Dimensionality-Driven Learning strategy (D2L) that modifies the loss function once the turning point between the two stages of dimensionality compression and expansion is recognized, in an effort to prevent overfitting.
  
  \item We empirically demonstrate on MNIST, SVHN, CIFAR-10 and CIFAR-100 datasets that our Dimensionality-Driven Learning strategy can effectively learn (1) low-dimensional representation subspaces that capture the underlying data distribution, (2) simpler hypotheses, and (3) high-quality deep representations.
\end{itemize}

\section{Related Work}
\label{sec:related}

\subsection{Generalization of DNNs}
\citeauthor{zhang2016understanding} (\citeyear{zhang2016understanding}) showed that DNNs are capable of memorizing completely random labels and exhibit poor generalization capability. They argued that DNNs employ case-by-case memorization on training samples and their labels in this scenario.
\citeauthor{krueger2017deep} (\citeyear{krueger2017deep}) highlighted that DNNs exhibit different learning styles on datasets with clean labels versus those on datasets with noisy inputs or noisy labels. They showed that DNNs require more capacity, longer training time to fit noisy labels and the learned hypothesis is more complex. \citeauthor{arpit2017closer} (\citeyear{arpit2017closer}) further substantiated this finding by identifying two stages of learning of DNNs with noisy labels: an early stage of simple pattern learning and refining, and a later stage of label memorization. They also showed that dropout regularization can hinder overfitting to noisy labels. \citeauthor{shwartz2017opening} (\citeyear{shwartz2017opening}) demonstrated that, on data with clean labels, DNNs with tanh layers undergo an initial label fitting phase and then a subsequent compression phase. They also argued that information compression is related to the excellent generalization performance of DNNs. However, \citeauthor{michael2018on} (\citeyear{michael2018on}) conducted experiments where information compression was not found to occur for ReLU \cite{glorot2011deep} DNNs.

While these works have studied the differences between learning with clean labels and learning with noisy labels, a full picture of this phenomenon and its implications for DNN generalization is yet to emerge.   Our study adds another perspective based on subspace dimensionality analysis, and shows how this can lead to the development of an effective learning strategy.   

\subsection{Noisy Label Learning}
A variety of approaches have been proposed to robustly train DNNs on datasets with noisy labels. One strategy is to explicitly or implicitly formulate the {\em noise model} and use a corresponding noise-aware approach. Symmetric label noise that is independent of the true label was modeled in \cite{larsen1998design}, and asymmetric label noise that is conditionally independent of individual samples was modeled in \cite{natarajan2013learning,sukhbaatar2014training}. There are also more complex noise models for training samples where true labels and noisy labels can be characterized by directed graphical models \cite{xiao2015learning}, conditional random fields \cite{vahdat2017toward}, neural networks \cite{veit2017learning,jiang2017mentornet} or knowledge graphs \cite{li2017learning}. These methods aim to correct noisy labels to their true labels via a clean label inference step or by assigning smaller weights to noisy label samples. For the modeling of label noise, they often require an extra dataset with ground truth of pre-identified noisy labels to be available, or an expensive detection process. They may also rely on specific assumptions about the noise model. 
Another approach is to use a refined training strategy that utilizes correction methods to adjust the loss function to eliminate the influence of noisy samples \citep{Wang_2018_CVPR}. Backward and Forward are two such correction methods that use an estimated or learned factor to modify the loss function \cite{patrini2017making}. A linear layer is added on top of the network to further augment the correction architecture in \cite{sukhbaatar2014learning,goldberger2016training}.  Bootstrap replaces the target labels with a combination of raw target labels and their predicted labels \cite{reed2014training}. 

Our proposed Dimensionality-Driven Learning strategy is also a loss correction method, one that avoids overfitting by using the estimation of the local intrinsic dimensionality of learned local subspaces to regulate the learning process. 
In Section \ref{sec:experiments} we empirically compare Dimensionality-Driven Learning with other loss correction strategies.

\subsection{Supervised Learning and Dimensionality}

The Local Intrinsic Dimensionality (LID) model \cite{houle2017local1} was recently used for successful detection of adversarial examples for DNNs by \cite{ma2018characterizing}. This work demonstrates that adversarial perturbations (one type of input noise) tend to increase the dimensionality of the local subspace immediately surrounding a test sample, and that features based on LID can be used for identifying such perturbations.
However, in this paper we show how LID can be used in a new way, as a tool for assessing the learning behavior of a DNN, and developing an adaptive learning strategy against noisy labels.

Other works have also considered the use of dimensionality measures for regularization in manifold learning \cite{roweis2000nonlinear,belkin2004regularization,belkin2006manifold}. For example, an intrinsic geometry regularization over Reproducing Kernel Hilbert Spaces (RKHS) was proposed in \cite{belkin2006manifold} to enforce smoothness of solutions relative to the underlying manifold, and a Laplacian-based regularization using the weighted neighborhood graph was proposed in \cite{belkin2004regularization}. In contrast to these works, which treated dimensionality as a characteristic of the global data distribution, we explore how knowledge of local dimensional characteristics can be used to monitor and modify DNN learning behavior for the noisy label scenario. 



\section{Dimensionality of Deep Representation Subspaces}
We now briefly introduce the LID measure for assessing the dimensionality of data subspaces residing in the deep representation space of DNNs. We then connect dimensionality theory with the learning process of DNNs.

\subsection{Local Intrinsic Dimensionality (LID)}
Local Intrinsic Dimensionality (LID) is an expansion-based measure of intrinsic dimensionality of the underlying data subspace/submanifold~\cite{houle2017local1}. 
In the theory of intrinsic dimensionality, classical
expansion models (such as the expansion dimension \cite{karger2002finding} and generalized expansion dimension \cite{HouleKN12}) measure the rate of growth in the number of data objects encountered as the distance from the reference sample increases. Intuitively, in Euclidean space, the volume of an $D$-dimensional ball grows proportionally to $r^D$ when its size is scaled by a factor of $r$. From the above rate of volume growth with distance, the dimension $D$ can be deduced from two volume measurements as: 
\begin{equation}
V_2/V_1 = (r_2/r_1)^D \Rightarrow D = \ln(V_2/V_1)/\ln(r_2/r_1).   
\end{equation}
The aforementioned expansion-based measures of intrinsic dimensionality would determine $D$ by estimating the volumes in terms of the numbers of data points captured by the balls. Transferring the concept of expansion dimension from the Euclidean space to the statistical setting of continuous distance distributions, the notion of ball volume is replaced by the probability measure associated with the balls. This leads to the formal definition of LID \cite{houle2017local1}:


\begin{definition}[Local Intrinsic Dimensionality] \quad \\
Given a data sample $x \in X$, let $r>0$ be a random variable denoting the distance from $x$ to other data samples. If the cumulative distribution function $F(r)$ is positive and continuously differentiable at distance $r>0$, the LID of $x$ at distance $r$ is given by:
\begin{equation} \label{eq:LID_r}
  \begin{split}
    \textup{LID}_F(r) & \triangleq \lim_{\epsilon\to 0} \frac{\ln\big(F((1+\epsilon) r)\big/F(r)\big)}{\ln(1+\epsilon)} 
     = \frac{r F'(r)}{F(r)},
  \end{split}
\end{equation}
whenever the limit exists.
\label{def:lid}
The \textup{LID} at $x$ is in turn defined as the limit of the radius $r \to 0$: 
\begin{equation} \label{eq:LID}
    \textup{LID}_F = \lim_{r \to 0}  \textup{LID}_F(r).
\end{equation}
\end{definition}
$\text{LID}_F$ describes the relative rate at which its cumulative distance function $F(r)$ increases as the distance $r$ increases. In the ideal case where the data in the vicinity of $x$ are distributed uniformly within a local submanifold, $\text{LID}_F$ equals the dimension of the submanifold. Nevertheless, in more general cases, LID also provides a rough indication of the dimension of the submanifold containing $x$ that would best fit the data distribution in the vicinity of $x$. 
 We refer readers to \cite{houle2017local1,houle2017local2} for more details about LID.


\textbf{Estimation of LID:}
Given a reference sample point $x \sim \mathcal{P}$, where $\mathcal{P}$ represents a global data distribution, $\mathcal{P}$ induces a distribution of distances relative to $x$ --- each sample $x_{*}\sim \mathcal{P}$ being associated with the distance value $d(x,x_{*})$. With respect to a dataset $X$ drawn from $\mathcal{P}$, the smallest $k$ nearest neighbor distances from $x$ can be regarded as extreme events associated with the lower tail of the induced distance distribution. From the statistical theory of extreme values, the tails of continuous distance distributions can be seen to converge to the Generalized Pareto Distribution (GPD), a form of power-law distribution \cite{coles2001introduction,hill1975simple}.
Several estimators of LID were developed in \cite{amsaleg2015estimating,levina2005maximum}, of which the Maximum Likelihood Estimator (MLE) exhibited the best trade-off between statistical efficiency and complexity:
\begin{equation} \label{eq:estimator}
\widehat{\textup{LID}}(x) = - \Bigg( \frac{1}{k}\sum_{i=1}^{k}\log \frac{r_i(x)}{r_{\mathit{max}}(x)}\Bigg)^{-1}.
\end{equation}
\noindent Here, $r_i(x)$ denotes the distance between $x$ and its $i$-th nearest neighbor, and $r_{\mathit{max}}(x)$ denotes the maximum of the neighbor distances. Note that the LID defined in Equation~\eqref{eq:LID} is a \textit{distributional} quantity, and the $\widehat{\textup{LID}}$ defined in Equation~\eqref{eq:estimator} is its \textit{estimate}.

\subsection{LID Estimation through Batch Sampling}\label{sec:lid_dnn}
Since computing neighborhoods with respect to the entire dataset $X$ can be prohibitively expensive, we will estimate LID of a training example $x$ from its $k$-nearest neighbor set within a \textit{batch} randomly selected from $X$.   
Consider a $L$-layer neural network $h:\mathcal{P} \rightarrow \mathbb{R}^c $, where $h^{(i)}$ is the intermediate transformation of the $i$-th layer, and $c$ is a positive number indicating the number of classes. Given a batch of training samples $X_{B}\subseteq X$, and a reference point $x \sim \mathcal{P}$ (not necessarily a training sample), we estimate the LID score of $x$ as:
\begin{equation} \label{eq:lid_training}
    \widehat{\textup{LID}}(x, X_B) = - \Bigg( \frac{1}{k}\sum_{i=1}^{k}\log \frac{r_i(g(x), g(X_{B}))}{r_{\mathit{max}}(g(x), g(X_{B}))}\Bigg)^{-1},
\end{equation}
where $g = h^{(L-1)}$ is the output of the second-to-last layer of the network, $r_i(g(x), g(X_{B}))$ is the distance of $g(x)$ to its $i$-th nearest neighbor in the transformed set $g(X_{B})$, and $r_{\mathit{max}}$ represents the radius of the neighborhood. $\widehat{\textup{LID}}(x, X_B)$ reveals the dimensional complexity of the {\em local subspace} in the vicinity of $x$, taken after transformation by $g$. Provided that the batch is chosen sufficiently large so as to ensure that the $k$-nearest neighbor sets remain in the vicinity of $g(x)$, the estimate of LID at $g(x)$ within the batch serves as an approximation to the value that would have been computed within the full dataset $g(X)$. 




\begin{figure}[!t]
\centering
\begin{subfigure}{.5\textwidth}
  \centering
  \includegraphics[width=0.48\linewidth]{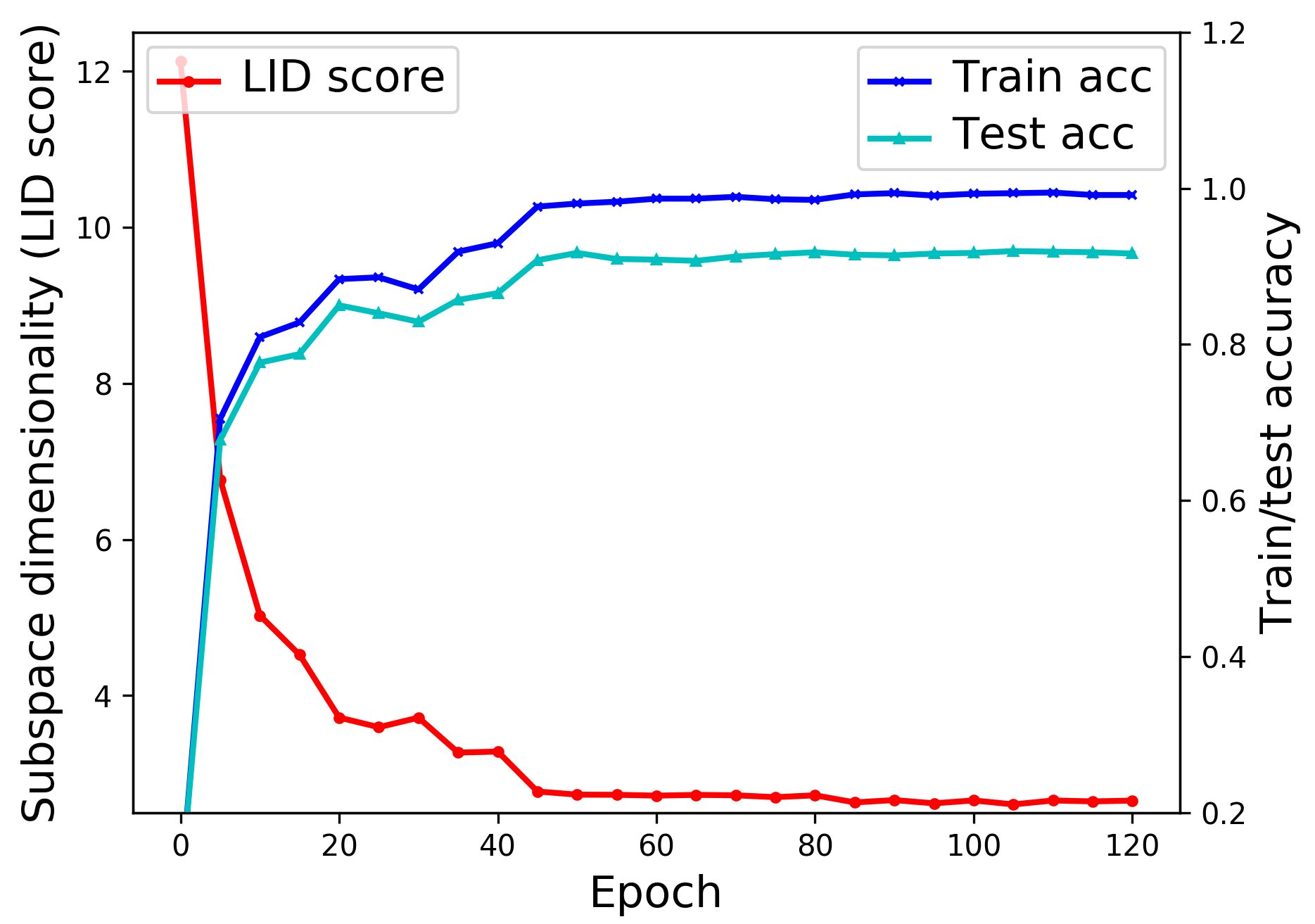}
\includegraphics[width=0.48\linewidth]{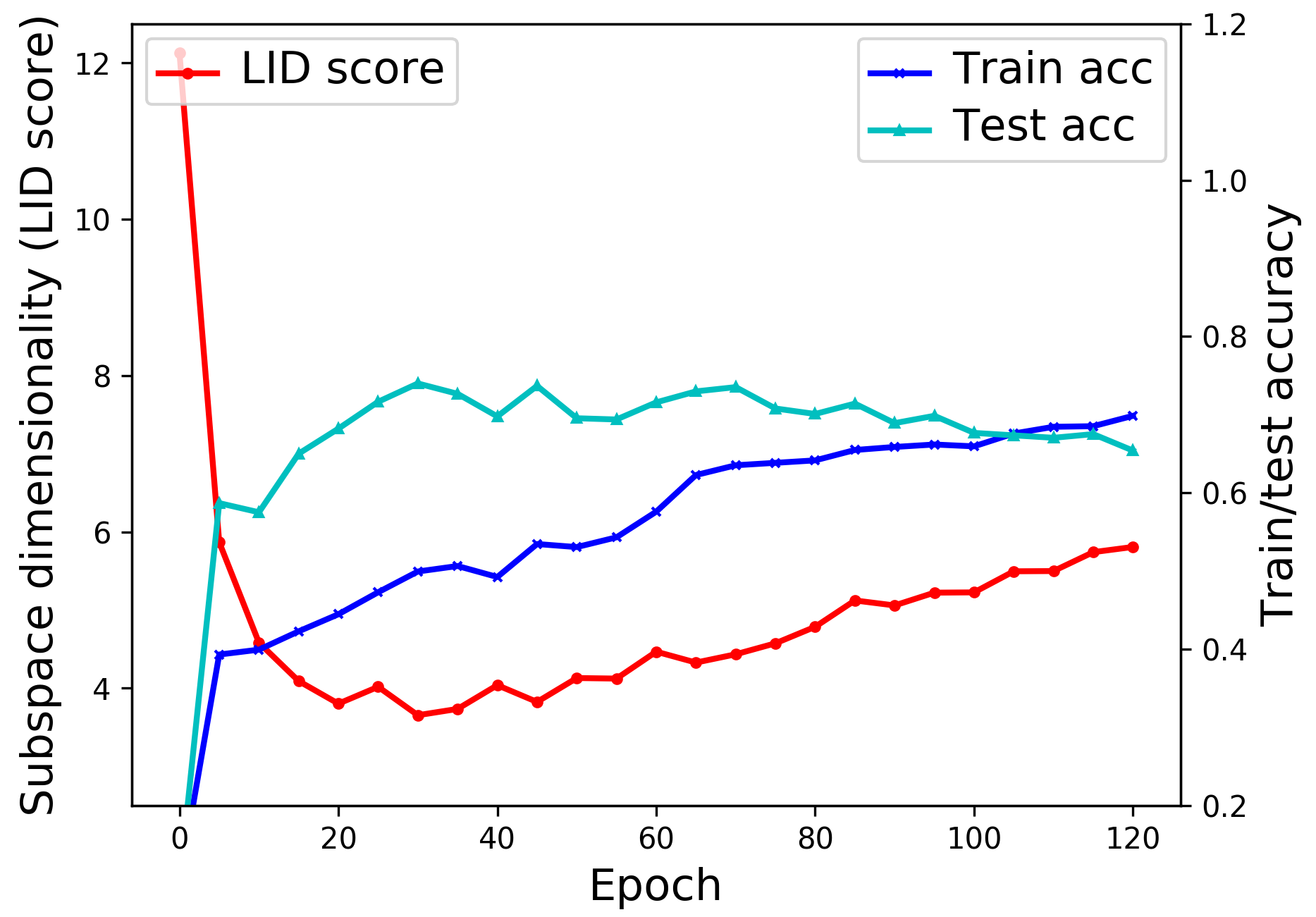}
  \caption{CIFAR-10}
  \label{fig:lid_cifar}
\end{subfigure} \\
\begin{subfigure}{.5\textwidth}
  \centering
  \includegraphics[width=0.48\linewidth]{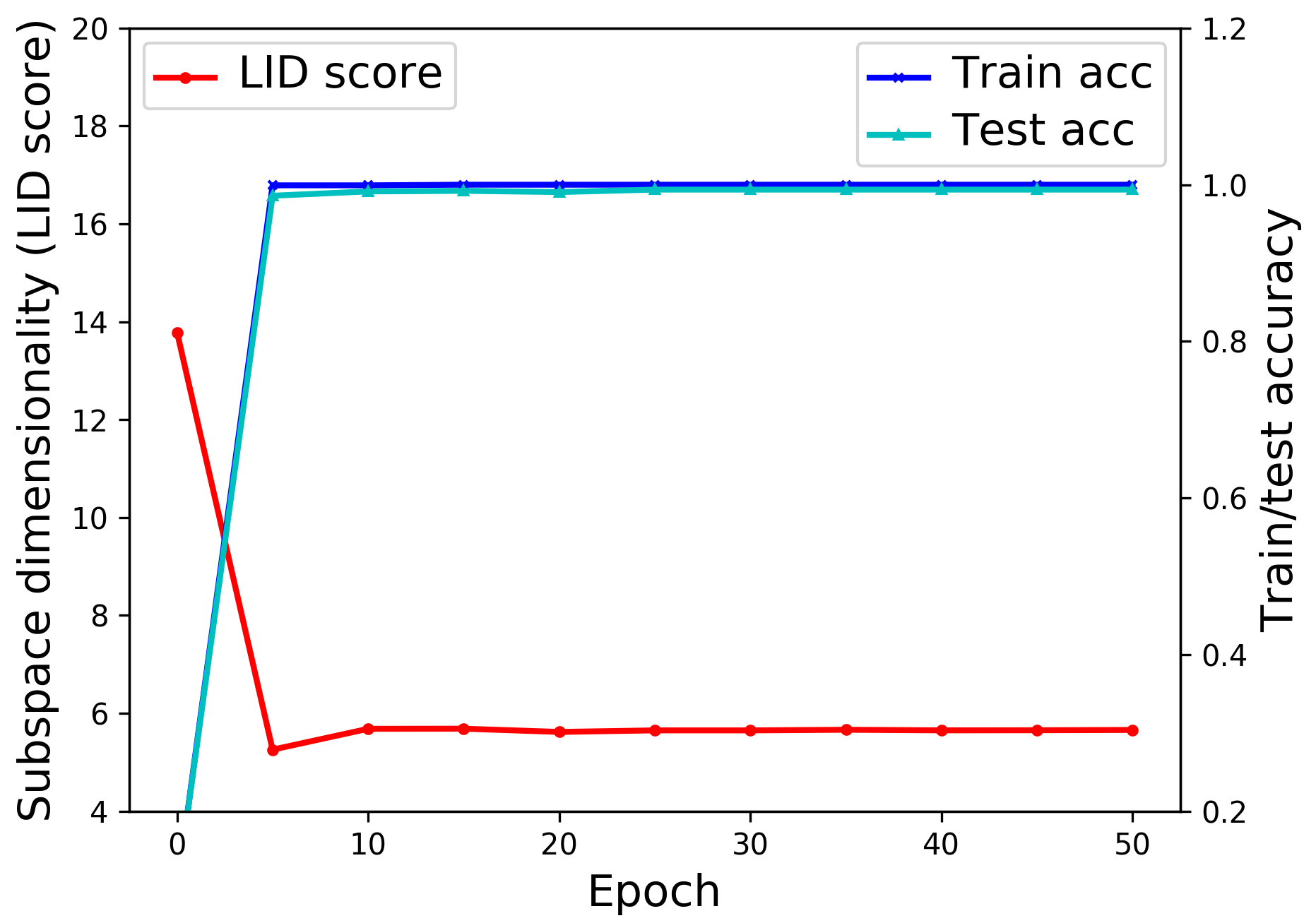}
\includegraphics[width=0.48\linewidth]{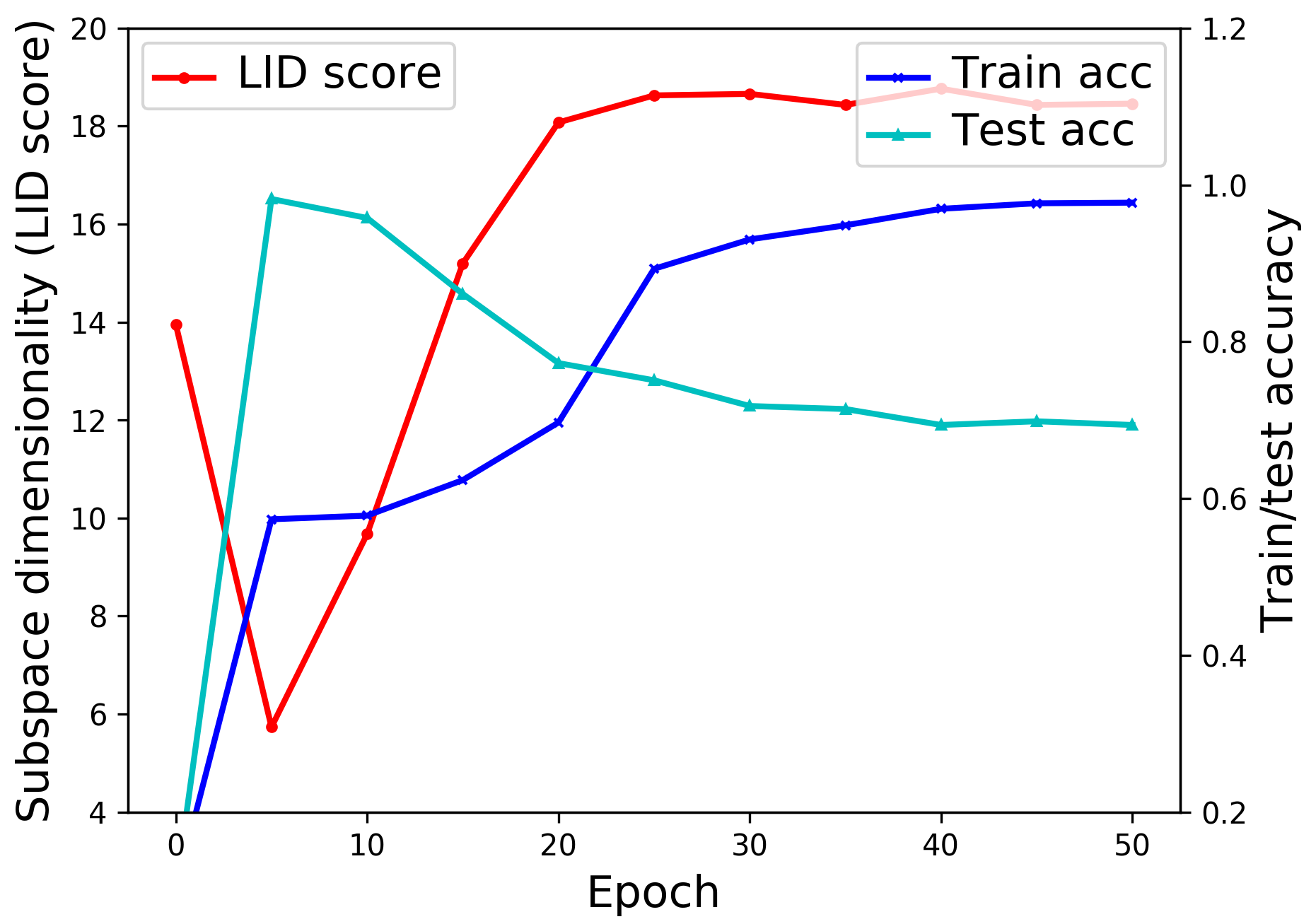}
  \caption{MNIST}
  \label{fig:lid_mnist}
\end{subfigure}
\caption{The subspace dimensionality (average LID scores) and train/test accuracy throughout training for a 12-layer CNN on CIFAR-10 (a) and a 5-layer CNN on MNIST (b) dataset with clean (left subfigures) and noisy labels (right subfigures). The average LID scores were computed at layer 11 for CIFAR-10 and layer 4 for MNIST.}
\label{fig:lid_trends}
\vspace{-0.2in}
\end{figure}

\subsection{Subspace Dimensionality and Noisy Labels}\label{sec:subsapce_dimensionality}
We now show by means of an example how the subspace dimensionality of training and test examples is affected by the quality of label information, as the number of training epochs is increased.
For our example, we trained a 5-layer Convolutional Neural Network (CNN) on MNIST (an image data set with 10 categories of handwritten digits \cite{lecun1998gradient}) and a 12-layer CNN on CIFAR-10 (a natural image data set with 10 categories \cite{krizhevsky2009learning}) using SGD, cross-entropy loss, and two different label quality settings: (1) clean labels for all training samples; (2) noisy labels for 40\% of the training samples, generated by uniformly and randomly replacing the correct label with one of the 9 incorrect labels. LID values at layer $4$ for MNIST and layer $11$ for CIFAR-10 were averaged over 10 batches of 128 points each, for a total of 1280 test points. The resulting LID scores and the train/test accuracies are shown in Figure \ref{fig:lid_trends}. When learning with clean labels, we observe a decreasing trend in LID score and an increasing trend in accuracy as the number of training epochs increases. However, when learning with noisy labels, we see a very different trend: first a decrease in LID followed by an increase, accompanied by an initial increase in test accuracy followed by a decrease. We observed similar dimensionality trends for a 6-layer CNN on SVHN \cite{netzer2011reading} and a 44-layer ResNet \cite{he2016deep} on CIFAR-100 \cite{krizhevsky2009learning}.

Clearly, in these two situations, the DNNs are exhibiting different learning styles. For training data with clean labels, the network gradually transforms the data to subspaces of low dimensionality. Once the subspaces of the lowest dimensionality has been found, the network effectively stops learning: the test accuracy stabilizes at its highest level and the dimensionality stabilizes at its lowest. On the other hand, for training data with noisy labels, the network initially learns a transformation of the data to subspaces of lower dimensionality, although not as low as when training on data with clean labels. Thereafter, the network progressively attempts to accommodate noisy labels by increasing the subspace dimensionality. 

\subsection{Two-Stage of Learning of DNNs on Noisy Labels}
From the above empirical results, we find that DNNs follow two-stage of learning in the presence of label noise: 1) an early stage of {\em dimensionality compression}, in which the dimensionalities associated with the underlying data manifold are learned; and 2) a later stage of {\em dimensionality expansion}, in which the subspace dimensionalities steadily increase as the learning process overfits to the noisy data. 

One possible explanation for this phenomenon can be found in the effect of transformation on the neighborhood set of test points.
Given a training point $x\in X$, its initial spatial location (before learning) would relate to a low-dimensional local subspace determined by the underlying manifold (call this subspace $A$). 
Although the initial neighborhood of $x$ would likely contain many data points that are also close to manifold $A$, the LID estimate would not
necessarily be the exact dimension of $A$. LID reveals the growth characteristics of the distance distribution from $x$, which is influenced by --- but not equal to --- the dimension of the manifold to which $x$ is best associated.

As the learning process progresses, the manifold undergoes a transformation by which it progressively achieves a better fit to the training data. If $x$ is labeled correctly, and if many of its neighbors also have clean labels, the learning process can be expected to converge towards a local subspace of relatively low intrinsic dimensionality (as observed in the left-hand plot of Figure~\ref{fig:lid_trends}); however, it should be noted that the learning process still risks overfitting to the data, if carried out too long. With overfitting, the dimensionality of the local manifold would be expected to rise eventually.

If $x$ is incorrectly labeled, each epoch in the learning process progressively causes $x$ --- or more precisely, its transform (call it $x'$) --- to migrate to a new local subspace (call it $A'$) associated with members of the same label that was incorrectly applied to $x$. During this migration, the neighborhood of $x'$ tends to contain more and more points of $A'$ that share the same label as $x$, and fewer and fewer points from the original neighborhood in $A$. With respect to the points of $A'$, the mislabeled point $x'$ is spatially an outlier, since its coordinates relate to $A$ and not $A'$; thus, the presence of $x'$ forces the local subspace around it to become more high-dimensional in order to accommodate (or compress) it. This distortion results in a {\em dimensionality expansion} in the vicinity of $x'$ that would be expected to be reflected in LID estimates based at $x'$. Stopping the learning process earlier allows $x'$ to find its neighborhood in $A$ before the local subspace is corrupted by too many neighbors from $A'$, which thus leads to better learning of the true data distribution and improved generalization to test data.


This explanation of the effect of incorrect labeling in terms of local subspaces is consistent with the one recently given in \cite{ma2018characterizing} for the effect of adversarial perturbation on DNN classification. 
In this situation, rather than directly assigning an incorrect label to the test item while leaving its spatial coordinates unchanged, the adversary must instead attempt to move a test point into a region associated with an incorrect class by means of an antagonistic learning process. 
In both cases, regardless of how the test point is modified, the neighborhoods of the transformed points are affected in a similar manner: as the neighborhood membership evolves, the local intrinsic dimensionality can be expected to rise. The associated changes in LID estimates have been used as the basis for the effective detection of a wide variety of adversarial attacks \citep{ma2018characterizing}.
Recent theoretical work for adversarial perturbation in nearest-neighbor classification further supports the relationship between LID and local transformation of data, by showing that the magnitude of the perturbation required in order to subvert the classification diminishes as the local intrinsic dimensionality and data sample size grow \citep{advperturbWIFS17}.

\section{Dimensionality-Driven Learning Strategy}
\label{sec:manifold-learning}
In the previous section, we observed that learning in the presence of noisy labels has two stages: dimensional compression, followed by dimensional expansion. Motivated by these observations, we propose a Dimensionality-Driven Learning (D2L) strategy whose objective is to avoid the overfitting and loss of test accuracy associated with dimensional expansion. 

Given a training sample $x$, we denote its raw label as $y$ and its predicted label as $\widehat y$, where both $y$ and $\widehat y$ are `one-hot' indicator vectors. $(\widehat{\textup{LID}}_{0}, \cdots, \widehat{\textup{LID}}_{i}, \cdots,  \widehat{\textup{LID}}_{T})$ is a sequence of LID scores, where $\widehat{\textup{LID}}_{i}$ represents the LID score computed from the second-to-last DNN layer at the $i$-th training epoch ($T$ epochs in total). Each LID score is produced as follows.   $m$ batches of samples are randomly selected $X_B^1,\ldots,X_B^m$ and for each $X_B^i$ and each of its members $x$, $\widehat{\textup{LID}}(x, X_B^i)$ is computed.    This gives $m \times |X_B^i|$ LID estimates, which are then averaged to compute the LID score for the epoch (later, in the experiments, we use $m=10$ and $|X_B^i|=128$

To avoid dimensionality expansion during training with noisy labels, we propose to reduce the effect of noisy labels on learning the true data distribution using the following adaptive LID-corrected labels:
\begin{equation} \label{eq:new_label}
    y^{*} = \alpha_i y + (1 - \alpha_i) \widehat y,
\end{equation}
where $\alpha_i$ is a LID-based factor that updates at the $i$-th training epoch:
\begin{equation} \label{eq:pace}
   \alpha_{i} = \exp \Big(- \lambda \frac{\widehat{\textup{LID}}_{i}}{\min_{j=0}^{i-1} \widehat{\textup{LID}}_{j}}\Big),
\end{equation}
where $\lambda=i / T$ is a weighting that indicates decreasing confidence in the raw labels when the training proceeds to the dimensionality expansion stage (that is, when LID begins to increase). The training loss can then be refined as:
\begin{equation} \label{eq:lid_loss}
\mathcal{L} = - \frac{1}{N}\sum_{n = 1}^{N} \sum_{y_n^*} y_n^* \log P(y_n^*|x_n),
\end{equation}
where $N$ is the total number of training samples and $P(y_n^*|x_n)$ is the predicted class probability of $y_n^*$ given $x_n$. 

Interpreting Equations~\eqref{eq:new_label} - \eqref{eq:lid_loss}, we can regard D2L as a simulated annealing algorithm that attempts to find an optimal trade-off between subspace dimensionality and prediction performance. The role of $\alpha$ is an exponential decay factor that allows for interpolation between raw and predicted label assignments according to the degree of dimensional expansion observed over the learning history. Here, dimensional expansion is assessed in terms of the ratio of two average LID scores: the score observed at the current epoch, and the lowest score encountered at earlier epochs. As the learning enters the dimensional expansion stage, this ratio exceeds 1, and the exponential decay factor begins to favor the current predicted label. The complete D2L learning strategy is shown in Algorithm \ref{alg:manifold}. 
Note that the computational cost of LID estimation through batch sampling is low compared to the overall training time ($t_{LID}/t_{training} \approx 1 - 2\%$), as it requires only the pairwise distances within a few batches.


To identify the turning point between the two stages of learning, we employ an epoch window of size $w \in [1, T-1]$ so as to allow $w$ epochs of initialization for the network, and to reduce the variation of stochastic optimization. The turning point is flagged when the LID score of the current epoch is two standard deviations higher than the mean LID score of the $w$ preceding epochs, until which the D2L loss is equivalent to the cross-entropy loss (enforced by setting $\alpha$ equal to $1$). The epoch at which the turning point is identified can be regarded as the first epoch at which overfitting occurs; for this reason, we roll the model state back to that of the previous epoch, and begin the interpolation between the raw and predicted label assignments. 
Although we find in the experimental results of Section \ref{sec:experiments} that this strategy works consistently well for a variety of datasets, further variations upon this basic strategy may also be effective. The D2L code is available at \url{https://github.com/xingjunm/dimensionality-driven-learning}.

\begin{algorithm}[tb]
   \caption{Dimensionality-Driven Learning (D2L)}
   \label{alg:manifold}
\begin{algorithmic}
   \STATE {\bfseries Input:} dataset $X$, network $h(x)$, total epochs $T$, epoch window $w$, number of batches for LID estimation  $m$.
   \STATE {\bfseries Initialize:}
   epoch $i\leftarrow 0$, $\mathit{lids} \leftarrow []$, $\alpha_0 \leftarrow 1$, turning epoch $u \leftarrow {-1}$.
  \REPEAT
   \STATE Train $h(x)$ for one epoch.
   \STATE $\mathit{lid} \leftarrow 0$, $\lambda \leftarrow i/T$.
   \FOR{$j = 1$ {\bfseries to} $m$}
   \STATE Sample $X_B$ from X.
   \STATE $\mathit{lid} \leftarrow \mathit{lid} + \frac{1}{|X_B|}\sum_{k=1}^{|X_B|}\widehat{\textup{LID}}(x, X_B)$.
   \ENDFOR
   \STATE $\mathit{lids[i]} \leftarrow \mathit{lid} / m$.
   \IF{ $i \geq w$ {\bfseries and} $u = {-1}$ {\bfseries and} \\ $\mathit{lid} - \textup{mean}(\mathit{lids}[i-w:i-1]) > 2\cdot\textup{std}(\mathit{lids}[i-w:i-1])$}
   \STATE $u \leftarrow i - 1$. ~~~~~\# \textit{turning point found}
   \STATE Rollback $h(x)$ to the $u$-th epoch.
   \ENDIF
   \IF{$u > {-1}$}
   \STATE $\alpha_i = \exp \big({-\lambda}\cdot\mathit{lids}[i] / \textup{min}(\mathit{lids}[0:i-1])\big)$.
   \ELSE
   \STATE $\alpha_i = \alpha_0$
  \ENDIF
   \STATE  $y^{*} = \alpha_i y + (1 - \alpha_i) \widehat y$.
   \STATE Update loss to $\mathcal{L} = - \frac{1}{N}\sum_{n = 1}^{N} \sum_{y_n^*} y_n^*\log P(y_n^*|x_n)$.
   \STATE $i \leftarrow i+1$.
   \UNTIL{$i = T$ {\bfseries or} early stopping.}
\end{algorithmic}
\end{algorithm}

\section{Experiments}\label{sec:experiments}
We evaluate our proposed D2L learning strategy, comparing the performance of our model with state-of-the-art baselines for noisy label learning.

\subsection{Empirical Understanding of D2L}
\label{sec:understanding}
We first provide an empirical understanding of the proposed D2L learning strategy on subspace learning, hypothesis learning, representation learning and model analysis.

\textbf{Experimental Setup:}
The experiments were conducted on the benchmark dataset CIFAR-10 \cite{krizhevsky2009learning}. We used a 12-layer CNN architecture.
All networks were trained using SGD with momentum 0.9, weight decay $10^{-4}$ and an initial learning rate of 0.1. The learning rate was divided by 10 after epochs 40 and 80 ($T=120$ epochs in total). 
Simple data augmentations (width/height shift and horizontal flip) were applied. Noisy labels were generated by introducing symmetric noise, in which the labels of a given proportion of training samples are flipped to one of the other class label, selected with equal probability. In \cite{vahdat2017toward} this noisy label generation scheme has been verified to be more challenging than that of restricted (asymmetric) label noise, which assumes that mislabelling only occurs within a specific set of classes \cite{reed2014training,patrini2017making}. 

\begin{figure}[!tb]
\centering
\begin{subfigure}{.5\textwidth}
  \centering
  \includegraphics[width=0.48\linewidth]{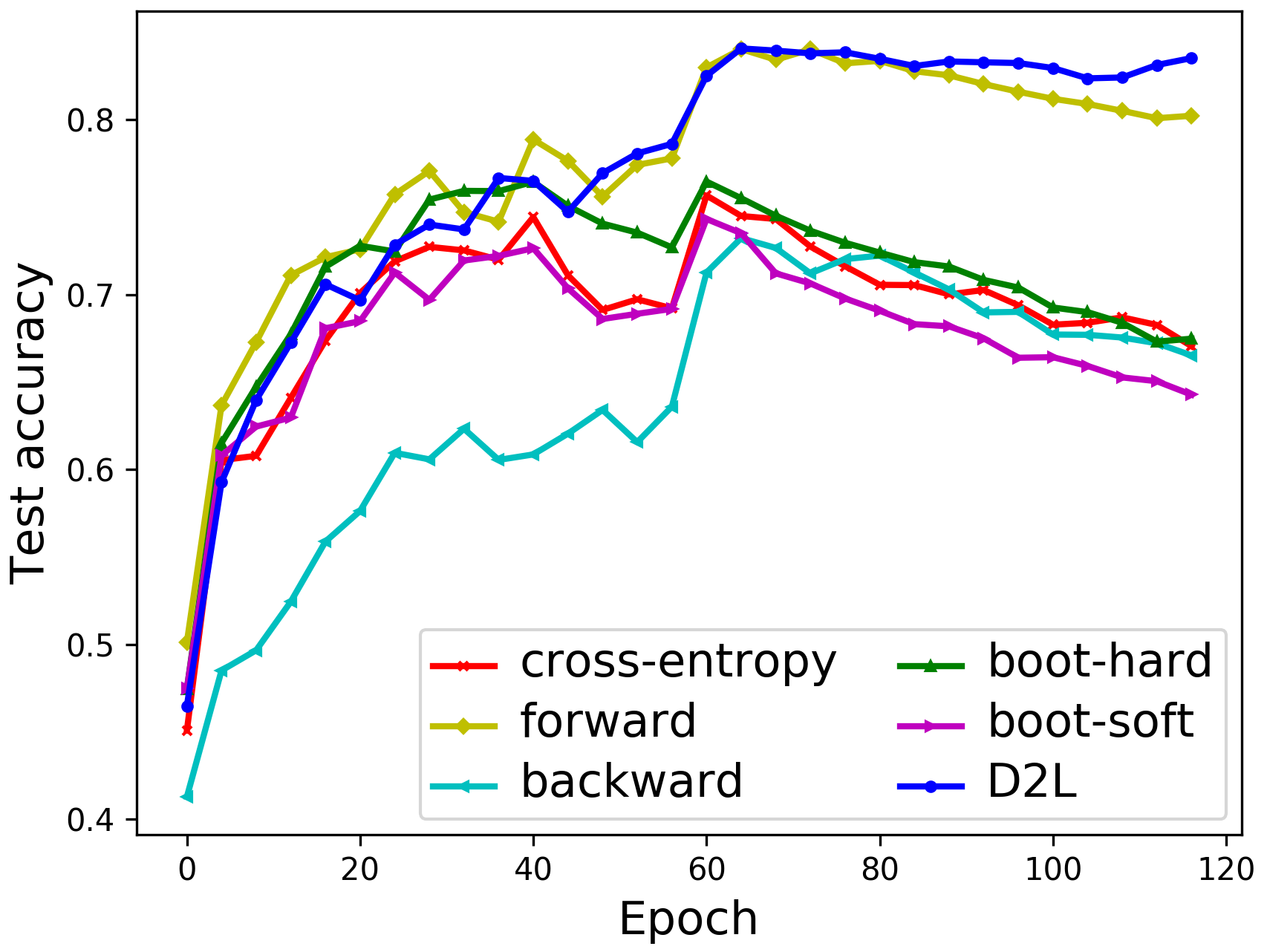}
\includegraphics[width=0.48\linewidth]{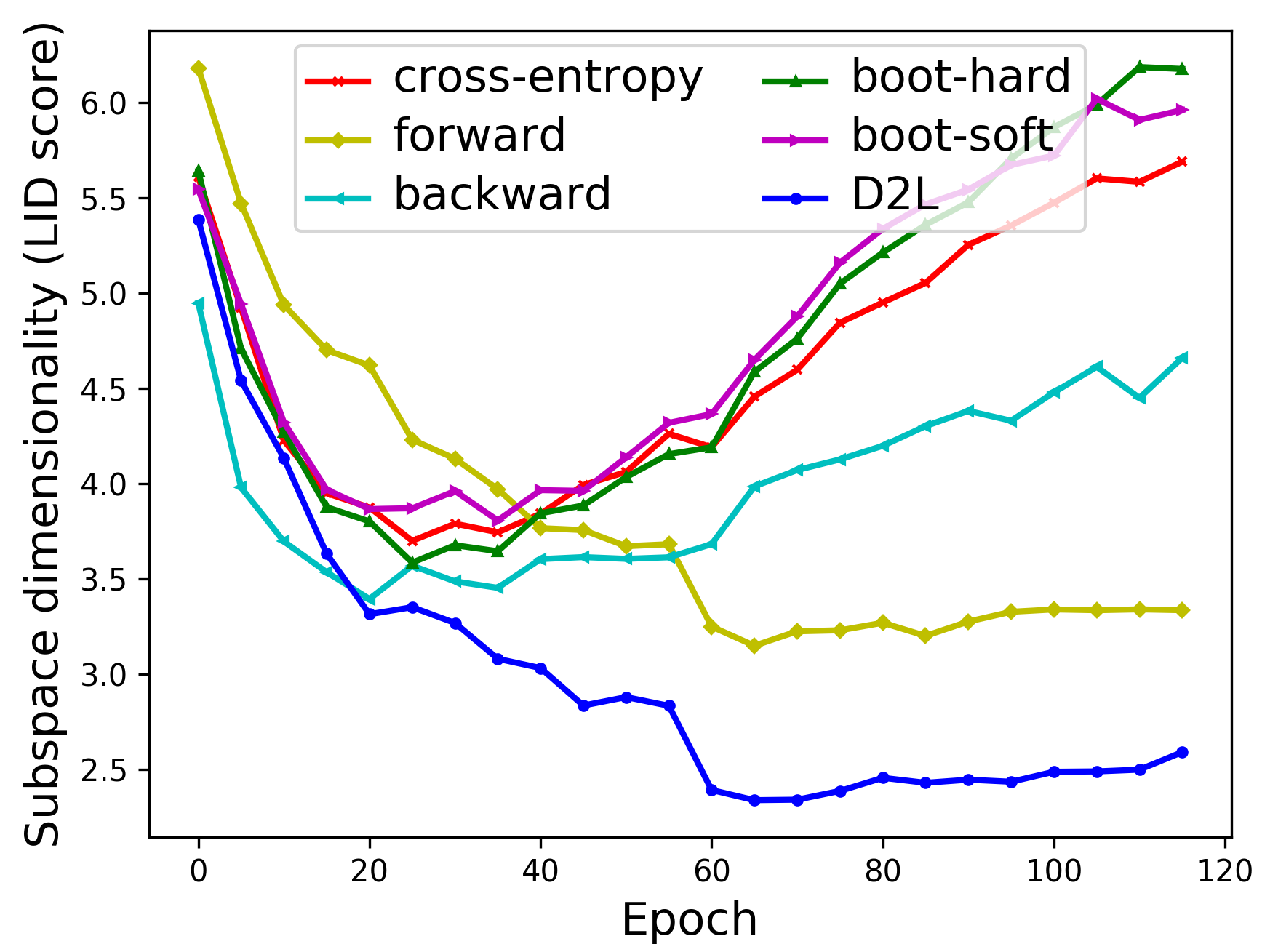}
  \caption{CIFAR-10 with 40\% noisy labels.}
  \label{fig:dimensionality_1}
\end{subfigure} \\
\begin{subfigure}{.5\textwidth}
  \centering
  \includegraphics[width=0.48\linewidth]{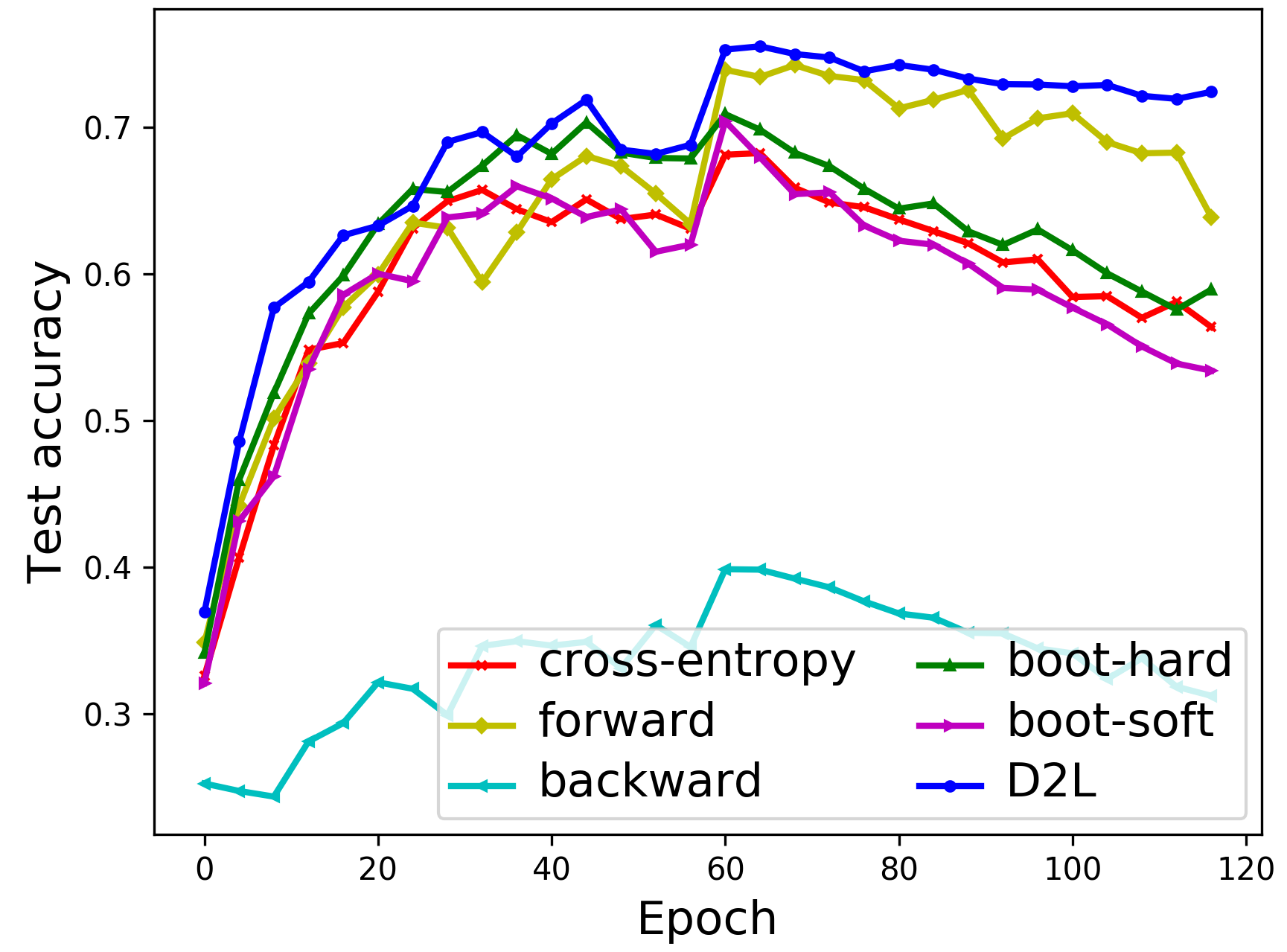}
\includegraphics[width=0.48\linewidth]{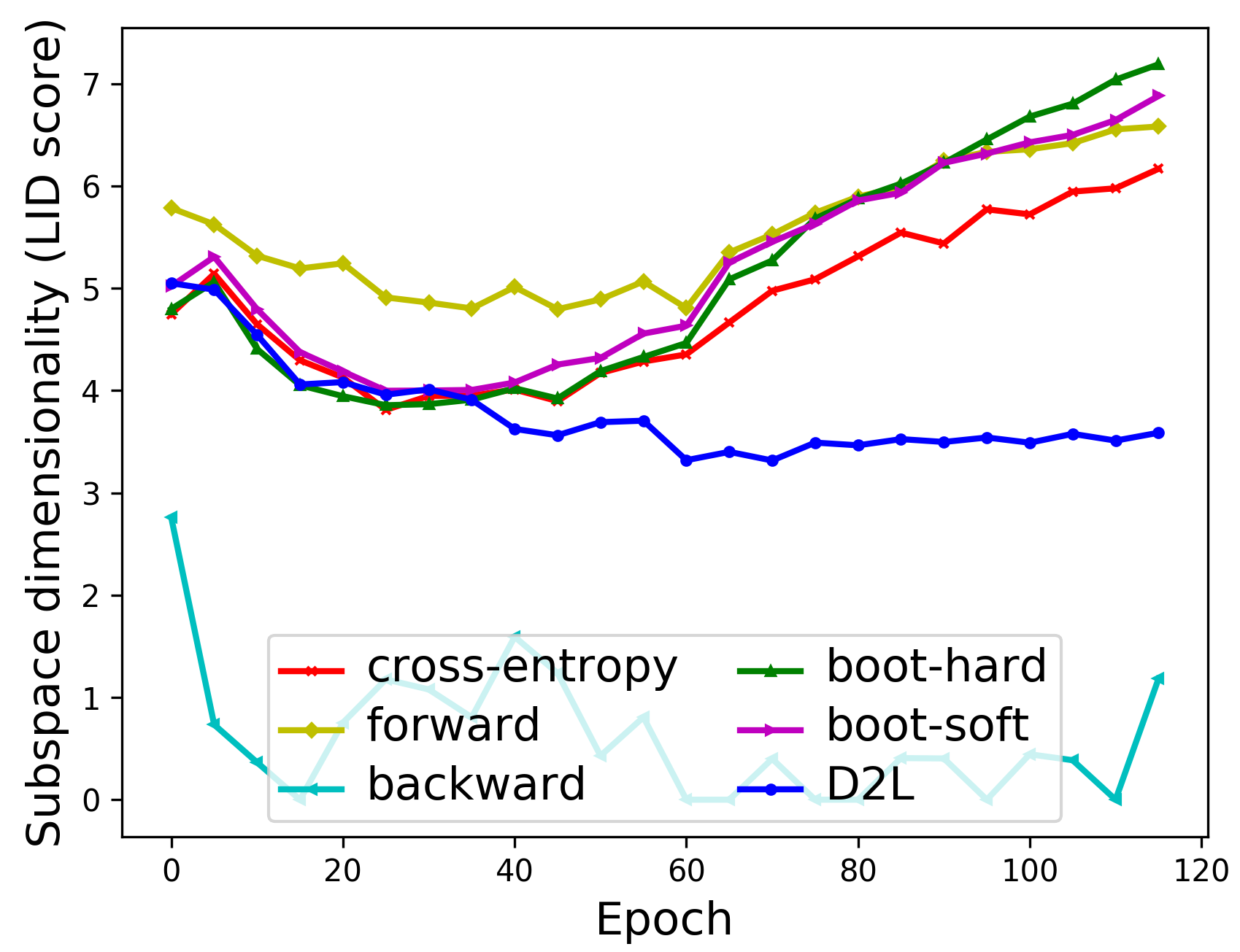}
  \caption{CIFAR-10 with 60\% noisy labels.}
  \label{fig:dimensionality_2}
\end{subfigure}
\caption{The trend of test accuracy and subspace dimensionality on CIFAR-10 with 40\% and 60\% noisy labels.}
\label{fig:dimensionality}
\vspace{-0.2in}
\end{figure}



\textbf{Competing Strategies:}
1) Backward \cite{patrini2017making}: training via loss correction by multiplying the cross-entropy loss by a noise-aware correction matrix; 2) Forward \cite{patrini2017making}: training with label correction by multiplying the network prediction by a noise-aware correction matrix; 3) Boot-hard \cite{reed2014training}: training with new labels generated by a convex combination (the ``hard" version) of the noisy labels and their predicted labels; 4) Boot-soft \cite{reed2014training}: training with new labels generated by a convex combination (the ``soft" version) of the noisy labels and their predictions; and 5) Cross-entropy: the conventional approach of training with cross-entropy loss.


The parameters of the competitors were configured according to their original papers. For our proposed D2L, we set $k=20$ for LID estimation, and used the average LID score over $m=10$ random batches of training samples as the overall dimensionality of the representation subspaces.


\textbf{Effect on Subspace Learning:}
We illustrate the effect of D2L on subspace learning by investigating the dimensionality (measured by LID) of the deep representation subspaces learned by DNNs and the test accuracy throughout training. The results are presented in Figure \ref{fig:dimensionality} for the CIFAR-10 dataset, with noisy label proportions set to 40\% and to 60\%. First, examining the test accuracy (the left-hand plots), we see that D2L can stabilize the test accuracy after around 60 epochs regardless of the noise rate, whereas the competitors experience a substantial decrease in test accuracy. This indicates the effectiveness of D2L in limiting the overfitting to noisy labels. Second, we focus on the dimensionality of the representation subspaces learned by different models (the right-hand plots). We observe that D2L is capable of learning representation subspaces which have significantly lower dimensionality than other models. It can also be noted that lower-dimensional subspaces lead to better generalization and higher test accuracy. This supports our claim that the true data distribution is of low dimensionality, and that D2L is capable of learning the low-dimensional true data distribution even with a large proportion of noisy labels. Note that for the case of 60\% label noise, the low test accuracy of the `backward' model, as well as the low dimensionality of the learned subspaces, together show that this competitor suffered from underfitting.


\begin{figure}[!ht]
\centering
\small
\includegraphics[width=0.48\linewidth]{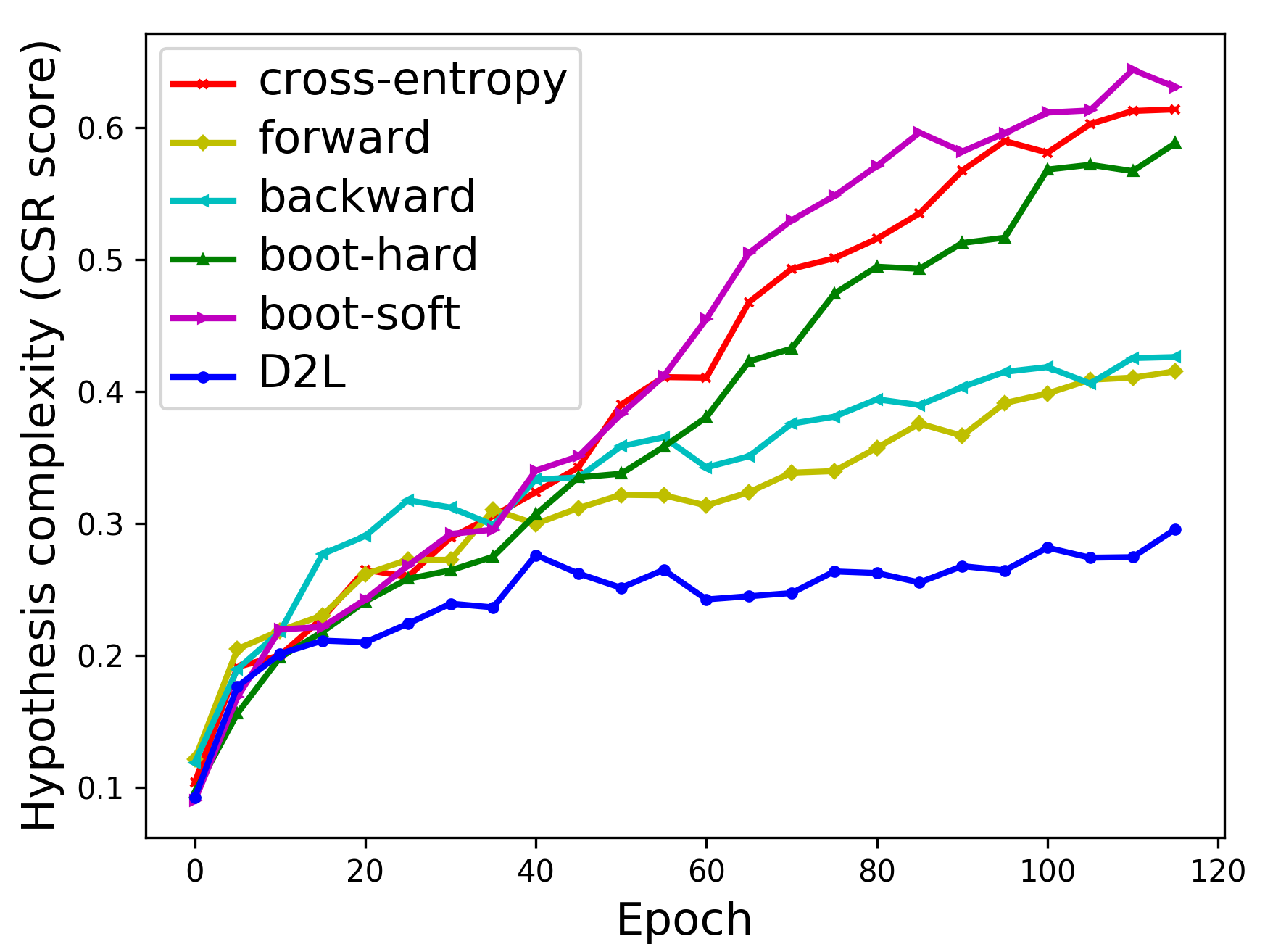}
\includegraphics[width=0.48\linewidth]{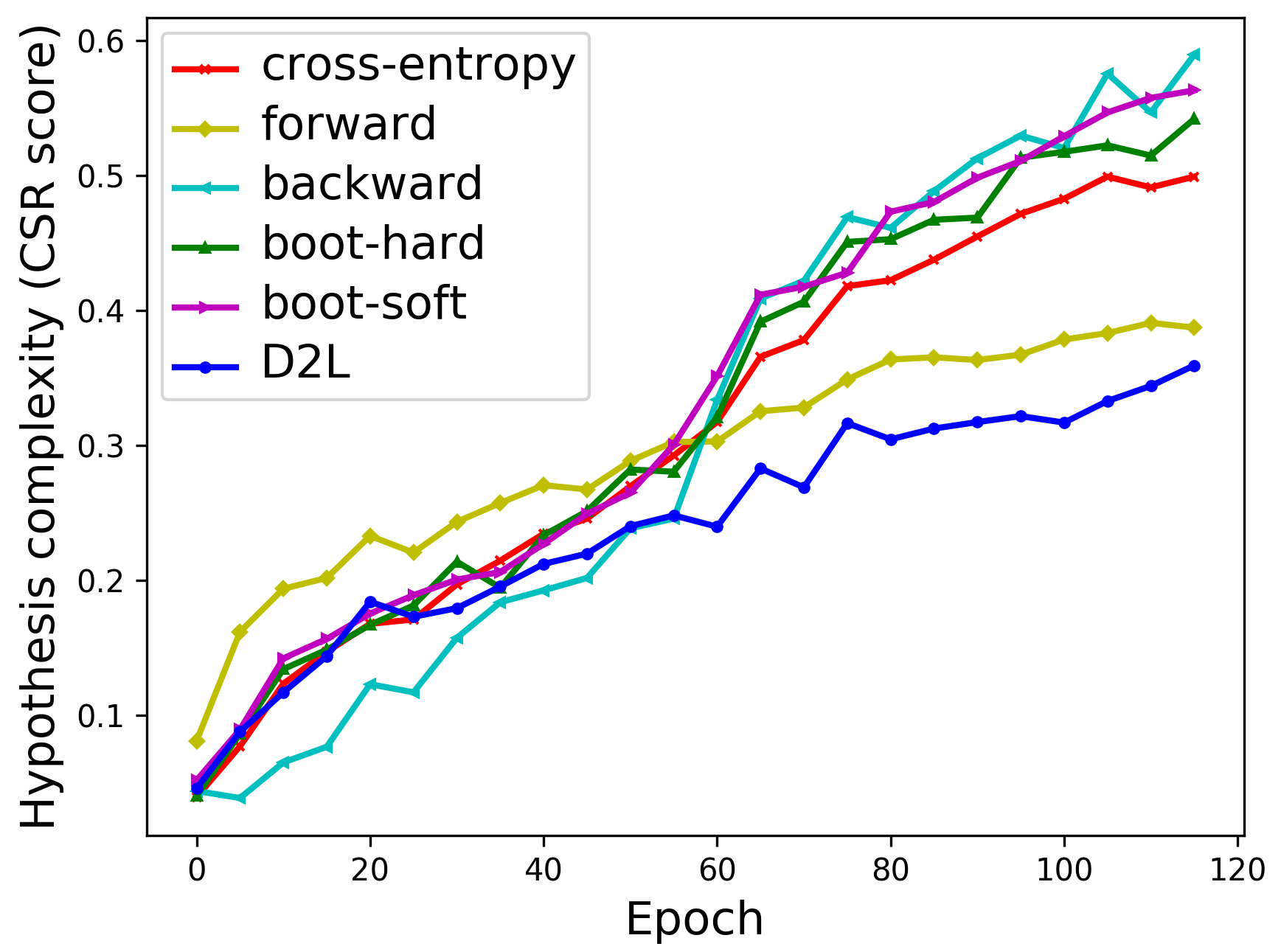}
\caption{The hypothesis complexity (measured by CSR) on CIFAR-10 with 40\% (left) and 60\% (right) noisy labels.}
\label{fig:complexity}
\end{figure}

\begin{figure}[!t]
\centering
\small
\includegraphics[width=0.98\linewidth]{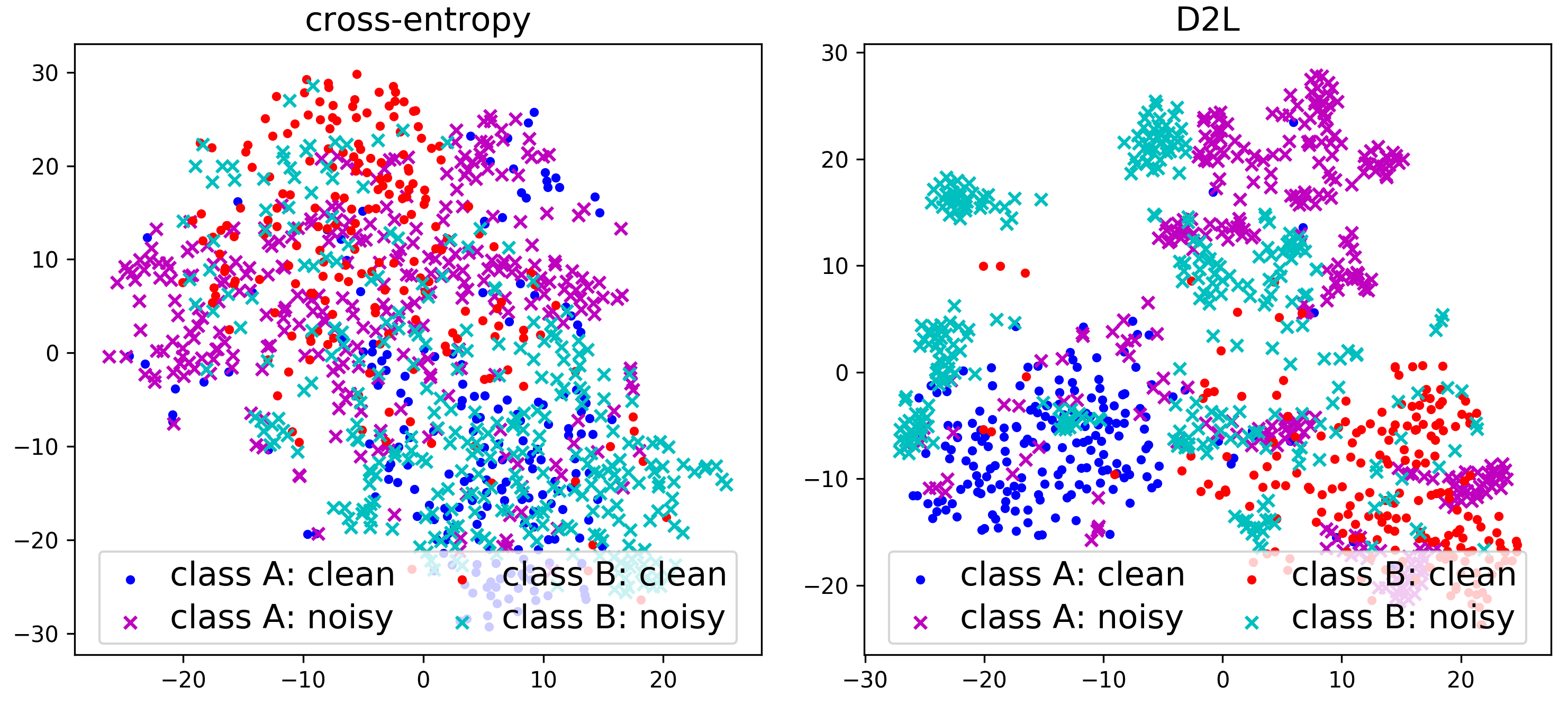}
\caption{Representations (t-SNE 2D embeddings) of two CIFAR-10 classes, `airplane' (A) and `cat' (B), learned by cross-entropy (left) and our D2L model (right), with 60\% of the class labels set to noise.}
\label{fig:representation}
\vspace{-0.2in}
\end{figure}

\textbf{Effect on Hypothesis Learning:}
\label{sec:complexity}
We investigate the complexity of the hypotheses learned from different models. Given a hypothesis space $\mathcal{H}$, a learned hypothesis $h \in \mathcal{H}$ from a DNN with lower complexity is expected to generalize better. 
Here, we use the recently proposed Critical Sample Ratio (CSR) \cite{arpit2017closer} as the measure for hypothesis complexity. CSR measures the density around the decision boundaries, where a high CSR score indicates a complex decision boundary and hypothesis.

As shown in Figure \ref{fig:complexity}, the complexity of the learned hypothesis from D2L is significantly lower than that of its competitors. Recalling the results from Figure \ref{fig:dimensionality}, where D2L achieved the highest test accuracy, we conclude that a simpler hypothesis does lead to better generalization, and that D2L is capable here of learning smoother decision boundaries and a simpler hypothesis than its competitors.


\begin{table*}[!tb]
\renewcommand{\arraystretch}{1.1}
\centering
\small
\caption{Test accuracy (\%) of different models on MNIST, SVHN, CIFAR-10 and CIFAR-100 with varying noise rates ($0\%$ -- $60\%$). The mean accuracy ($\pm$std) over 5 repetitions of the experiments are reported, and the best results are highlighted in \textbf{bold}.}
\label{tb:experiment_sym}
\begin{tabular}{ll|cccccc}
\hline
\multicolumn{2}{c|}{Dataset / Noise Rate} & cross-entropy & forward & backward & boot-hard & boot-soft & D2L  \\ \hline
\multirow{4}{*}{MNIST} 
& 0\% & 99.24$\pm$0.0 & \textbf{99.30$\pm$0.0} & 99.23$\pm$0.1 & 99.13$\pm$0.2 & 99.20$\pm$0.0 & 99.28$\pm$0.0  \\
& 20\% & 88.02$\pm$0.1 & 96.45$\pm$0.1 & 90.12$\pm$0.1 & 87.69$\pm$0.2 & 88.50$\pm$0.1 & \textbf{98.84$\pm$0.1}  \\
& 40\% & 68.46$\pm$0.1  & 94.90$\pm$0.1 & 70.89$\pm$0.1  & 69.49$\pm$0.2 & 70.19$\pm$0.2 & \textbf{98.49$\pm$0.1}  \\
& 60\% & 45.51$\pm$0.2 & 82.88$\pm$0.1 & 52.83$\pm$0.2 & 50.45$\pm$0.1 & 46.04$\pm$0.1 & \textbf{94.73$\pm$0.2}  \\ \hline
\multirow{4}{*}{SVHN}
& 0\% & 90.12$\pm$0.0 & 90.22$\pm$0.1 & 90.16$\pm$0.1 & 89.47$\pm$0.0 & 89.26$\pm$0.0 & \textbf{90.32$\pm$0.0} \\
& 20\% & 79.10$\pm$0.1 & 85.51$\pm$0.1 & 79.61$\pm$0.2 & 81.21$\pm$0.1 & 79.26$\pm$0.2 & \textbf{87.63$\pm$0.1}  \\
& 40\% & 62.92$\pm$0.1  & 79.09$\pm$0.2 & 64.15$\pm$0.1 & 63.25$\pm$0.2 & 64.30$\pm$0.2 & \textbf{82.68$\pm$0.1}  \\
& 60\% & 38.54$\pm$0.2 & 62.57$\pm$0.2 & 53.14$\pm$0.1 & 47.61$\pm$0.2 & 39.21$\pm$0.2 & \textbf{80.92$\pm$0.2}  \\ \hline
\multirow{4}{*}{CIFAR-10}
& 0\% & 89.31$\pm$0.1 & \textbf{90.27$\pm$0.1} & 89.03$\pm$0.2 & 89.06$\pm$0.3 & 89.46$\pm$0.2 & 89.41$\pm$0.2  \\
& 20\% & 81.52$\pm$0.1  & 84.61$\pm$0.3 & 79.41$\pm$0.1 & 81.19$\pm$0.4 & 79.21$\pm$0.2 & \textbf{85.13$\pm$0.2}  \\
& 40\% & 73.51$\pm$0.3  & 82.84$\pm$0.2 & 74.69$\pm$0.2 & 76.67$\pm$0.2 & 73.81$\pm$0.1 & \textbf{83.36$\pm$0.3}  \\
& 60\% & 67.03$\pm$0.3  & 72.41$\pm$0.4 & 45.42$\pm$0.4 & 70.57$\pm$0.3 & 68.12$\pm$0.2 & \textbf{72.84$\pm$0.3}  \\
\hline
\multirow{4}{*}{CIFAR-100}
& 0\% & 68.20$\pm$0.2 & 68.54$\pm$0.3 & 68.48$\pm$0.3 & 68.31$\pm$0.2 & 67.89$\pm$0.2 & \textbf{68.60$\pm$0.3}  \\
& 20\% & 52.88$\pm$0.2  & 60.25$\pm$0.2 & 58.74$\pm$0.3 & 58.49$\pm$0.4 & 57.32$\pm$0.3 & \textbf{62.20$\pm$0.4}  \\
& 40\% & 42.85$\pm$0.2  & 51.27$\pm$0.3 & 45.42$\pm$0.2 & 44.41$\pm$0.1 & 41.87$\pm$0.1 & \textbf{52.01$\pm$0.3}  \\
& 60\% & 30.09$\pm$0.2  & 41.22$\pm$0.3 & 34.49$\pm$0.2 & 36.65$\pm$0.3 & 32.29$\pm$0.1 & \textbf{42.27$\pm$0.2}  \\
\hline
\end{tabular}
\end{table*}

\textbf{Effect on Representation Learning:}
\label{sec:regularization}
To analyze the effectiveness of D2L for representation learning, we visualize dataset representations in 2-dimensional embeddings using t-SNE \cite{maaten2008visualizing}, a commonly-used dimensionality reduction technique for the visualization of high-dimensional data \cite{lecun2015deep}. Figure \ref{fig:representation} presents the reduced 2D embeddings of 500 randomly selected samples from each of two classes on CIFAR-10. For each class, 40\% of the samples were assigned correct labels (the `clean' samples), and 60\% were assigned incorrect labels chosen uniformly at random from the 9 other classes (the `noisy' samples). We see that D2L (the right-hand plot) can learn high-quality representations that accurately separate the two classes of objects (blue vs red), and can effectively isolate noisy samples (magenta/cyan) from clean samples (blue/red). However, for both classes, representations learned by cross-entropy (the left-hand plot) suffer from significant overlapping between clean and noisy samples. Note that the representations of noisy samples learned by D2L are more fragmented, since the noisy labels are from many different classes. Overall, D2L is able to learn a high-quality representation from noisy datasets.

\begin{figure}[!t]
\centering
\small
\includegraphics[width=0.48\linewidth]{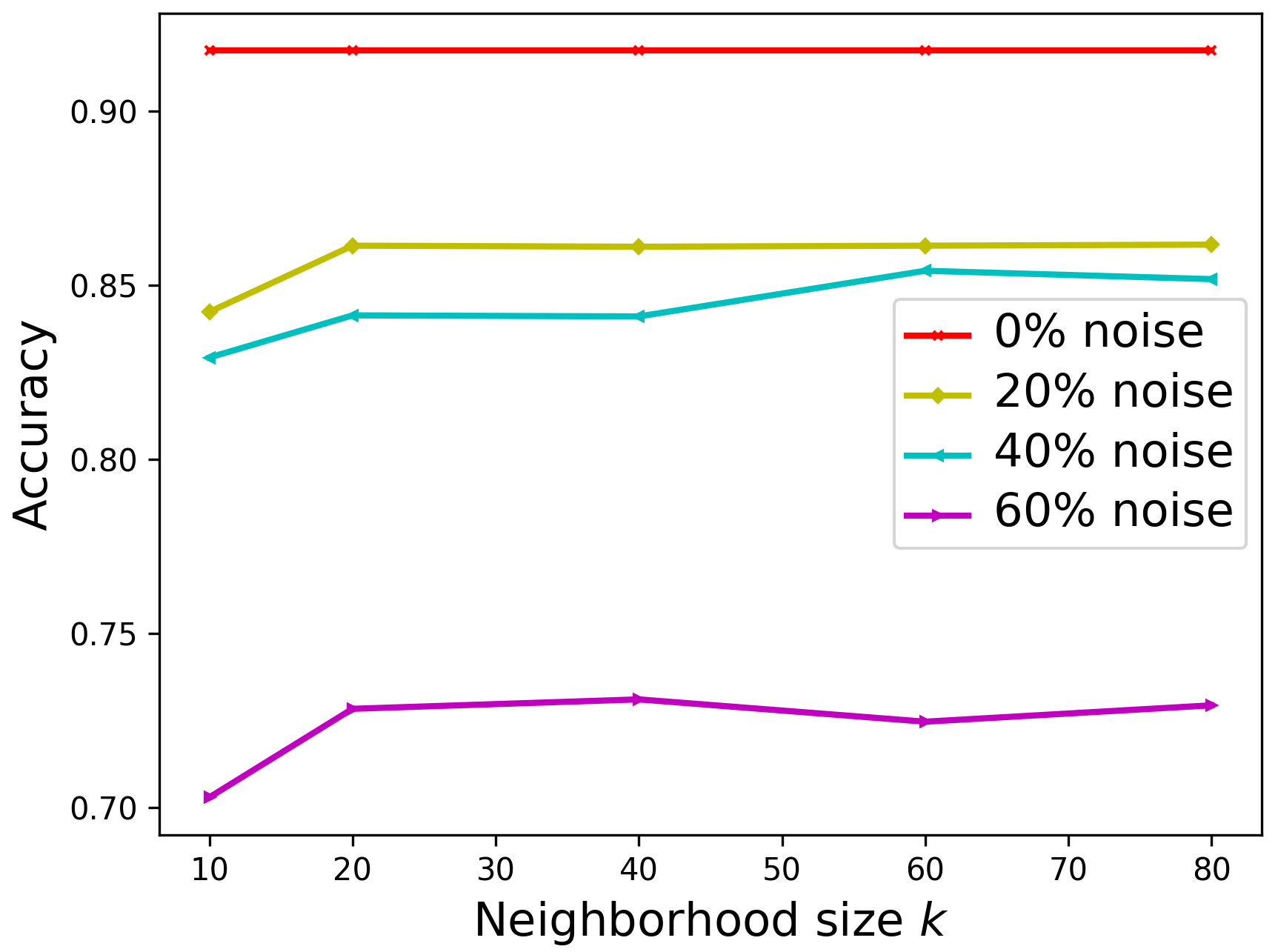}
\includegraphics[width=0.48\linewidth]{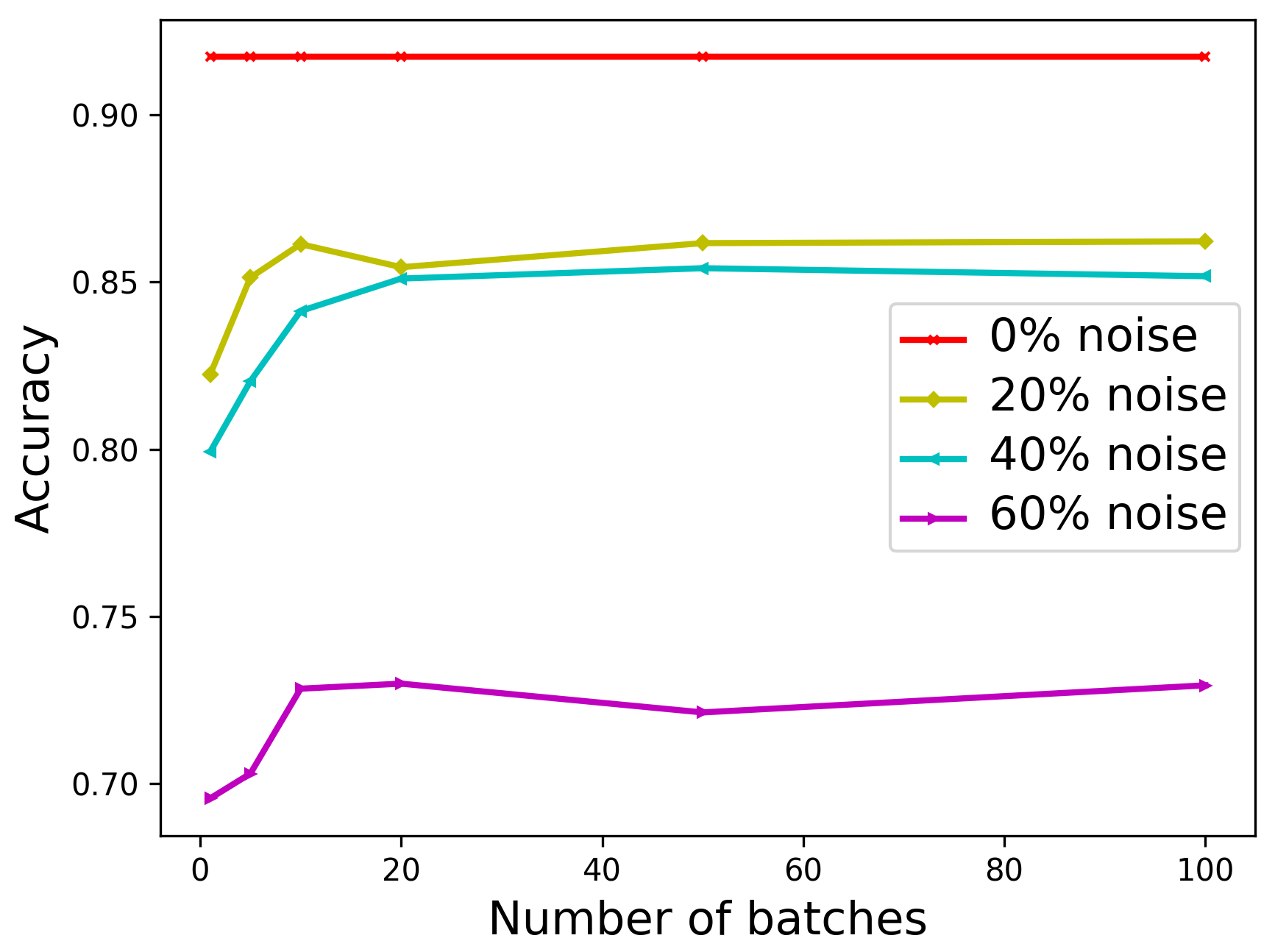}
\caption{Grid searching neighborhood size $k$ (left) and number of batches $m$ (right) for the estimation of LID on CIFAR-10 with various noise rate.}
\label{fig:tuning_k}
\vspace{-0.2in}
\end{figure}

\textbf{Parameter Sensitivity:} We assess the sensitivity of D2L to the neighborhood size $k$ and the number of batches $m$ used to compute the mean LID. Figure \ref{fig:tuning_k} shows that D2L is relatively insensitive to these two hyper-parameters on the CIFAR-10 dataset. We observed similar behavior with the other three datasets.

\subsection{Robustness against Noisy Labels}\label{sec:robustness_test}
Finally, we evaluate the robustness of D2L against noisy labels under varying noise rates (0\%, 20\%, 40\%, and 60\%) on several benchmark datasets, comparing to state-of-the-art baselines for noisy label learning.

\textbf{Experimental Setup:} Experiments were conducted on several benchmark datasets: MNIST \cite{lecun1998gradient}, SVHN \cite{netzer2011reading}, CIFAR-10 \cite{krizhevsky2009learning} and CIFAR-100 \cite{krizhevsky2009learning}. We used a LeNet-5 network \cite{lecun1998gradient} for MNIST, a 6-layer CNN for SVHN, a 12-layer CNN for CIFAR-10 and a ResNet-44 network \cite{he2016deep} for CIFAR-100. 
All networks were trained using SGD with momentum 0.9, weight decay $10^{-4}$ and an initial learning rate of 0.1. The learning rate is divided by 10 after epochs 20 and 40 for MNIST/SVHN (50 epochs in total), after epochs 40 and 80 for CIFAR-10 (120 epochs in total), and after epochs 80, 120 and 160 for CIFAR-100 (200 epochs in total) \cite{huang2016deep}. 
Simple data augmentations (width/height shift and horizontal flip) were applied on CIFAR-10 and CIFAR-100. Noisy labels were generated as described in Section~\ref{sec:understanding}. On a particular dataset, the compared methods differ only in their loss functions --- they share the same CNN architecture, regularizations (batch normalization and max pooling), and the number of training epochs. We repeated the experiments 5 times with different random seeds for network initialization and label noise generation.

\textbf{Results:} We report the mean test accuracy and standard deviation over 5 repetitions of the experiments
in Table \ref{tb:experiment_sym}. D2L outperforms its competitors consistently across all datasets and across all noise rates tested. In particular, the performance gap between D2L and its competitors increases as the noise rate is increased from 20\% to 60\%. We also note that as the noise rate increases, the accuracy drop of D2L is the smallest among all models. Even with 60\% label noise, D2L can still obtain a relatively high classification accuracy, which indicates that D2L may have the potential to be an effective strategy for semi-supervised learning. 


\section{Discussion and Conclusion}
In this paper, we have investigated the generalization behavior of DNNs for noisy labels in terms of the intrinsic dimensionality of local subspaces. We observed that dimensional compression occurs early in the learning process, followed by dimensional expansion as the process begins to overfit. Employing a simple measure of local intrinsic dimensionality (LID), we proposed a Dimensionality-Driven Learning (D2L) strategy for avoiding overfitting that identifies the learning epoch at which the transition from dimensional compression to dimensional expansion occurs, and then suppresses the subsequent dimensionality expansion. D2L delivers very strong classification performance across a range of scenarios with high proportions of noisy labels.   

We believe that dimensionality-based analysis opens up new directions for understanding and enhancing the behavior of DNNs.
Theoretical formulation of DNN subspace dimensionality, and investigation of the effects of data augmentation and regularization techniques such as batch normalization \cite{ioffe2015batch} and dropout \cite{srivastava2014dropout} are possible directions for future research.  
%
Another open issue is the investigation of how other forms of noise such as adversarial or corrupted inputs and asymmetric label noise
can affect local subspace dimensionality and DNN learning behavior.



\section*{Acknowledgements}
James Bailey is in part supported by the Australian Research Council via grant number DP170102472. Michael~E.~Houle is partially supported by JSPS Kakenhi Kiban (B) Research Grants 15H02753 and 18H03296. Shu-Tao Xia is partially supported by the National Natural Science Foundation of China under grant No.~61771273.
\bibliography{icml2018}
\bibliographystyle{icml2018}






\end{document}